# SIFT-based Ear Recognition by Fusion of Detected Keypoints from Color Similarity Slice Regions

Dakshina Ranjan Kisku, *Member, IEEE*, Hunny Mehrotra, Phalguni Gupta, and Jamuna Kanta Sing, *Member, IEEE*

*Abstract*—Ear biometric is considered as one of the most reliable and invariant biometrics characteristics in line with iris and fingerprint characteristics. In many cases, ear biometrics can be compared with face biometrics regarding many physiological and texture characteristics. In this paper, a robust and efficient ear recognition system is presented, which uses Scale Invariant Feature Transform (SIFT) as feature descriptor for structural representation of ear images. In order to make it more robust to user authentication, only the regions having color probabilities in a certain ranges are considered for invariant SIFT feature extraction, where the K-L divergence is used for keeping color consistency. Ear skin color model is formed by Gaussian mixture model and clustering the ear color pattern using vector quantization. Finally, K-L divergence is applied to the GMM framework for recording the color similarity in the specified ranges by comparing color similarity between a pair of reference model and probe ear images. After segmentation of ear images in some color slice regions, SIFT keypoints are extracted and an augmented vector of extracted SIFT features are created for matching, which is accomplished between a pair of reference model and probe ear images. The proposed technique has been tested on the IITK Ear database and the experimental results show improvements in recognition accuracy while invariant features are extracted from color slice regions to maintain the robustness of the system.

## I. INTRODUCTION

IN recent years, ear recognition [1] has emerged as one of the most reliable and effective biometrics authentication systems. However, there are very few biometrics characteristics available, which have already been proved to be secure means of authentication to several security and access control applications. Such as fingerprint [2], iris [3], palmprint [4] biometrics, etc. Although, thrust for more secure and robust biometric traits are still a major problem due to several implementation constraints. Among the existing biometrics traits, ear biometric [1], [5] considered as a most viable and robust biometric characteristic in line with iris, fingerprint biometrics. Ear biometric can be often compared with face biometrics [6], [7], [8] regarding many physiological aspects. Ears [1], [5] have several advantages over facial features [7], [8], such as uniform distributions of intensity and spatial resolution, and less variability with expressions and orientation of the face. Unlike face recognition [7], [8] with changing lightning, and different pose of head positions, ear shape does not changes over time and ageing. Further low effect of lighting conditions and spatial distribution of pixels made ear biometrics an emerging person authentication and access control system.

Although many 2D and 3D ear recognitions [9] have been proposed in the literatures so far, ear recognition remains a challengeable problem in biometrics in order to establish a robust person authentication. In any biometrics authentication [10] problem, feature extraction and representation are very important phases in personal recognition. Until and unless some distinct and invariant features have been extracted, higher accuracy can not be achieved. Therefore, the invariant features, which are extracted from higher matching probability texture regions, would be appropriate methodology for development of any biometrics system. In such cases, strong feature representation with invariance property and suitable selection of pattern classifier can make any biometric system [10] for deployment to commercial use.

Over the last few years, substantial improvements have done in biometrics authentications [10]. Performance priority based cutting-edge biometric technologies including ear recognition have attracted attentions due to invariant texture patterns and physiological structures of biometrics characteristics. Many researchers have presented and discussed ear biometrics as user recognition for both controlled/uncontrolled environments. In initial phase of object recognition, many pattern classification and recognition techniques with various feature representation techniques have been presented in [11], [12]. Due to thrust for robust object recognition many researchers have still faith on invariant features. As a result, some exciting 2D and 3D ear recognition techniques [5], [9], [13] have been proposed by using the general object recognition techniques [1], [9], [14]. In these techniques [1], [9], [14] ear recognitions have been developed under varying lighting conditions and poses with varying performances. Performances often degrade due to uncontrolled lighting

Dakshina Ranjan Kisku is with the Department of Computer Science and Engineering, Dr. B. C. Roy Engineering College, Durgapur – 713206, India (phone: +91-9732111234; fax: 303-555-5555; e-mail: drkisku@ ieee.org).
Hunny Mehrotra is with the Department of Computer Science and Engineering, National Institute of Technology Rourkela, Rourkela – 769008, India (e-mail: hunny@nitrkl.ac.in).
Phalguni Gupta is with the Department of Computer science and Engineering, Indian Institute of Technology Kanpur, Kanpur – 208016, India (e-mail: pg@cse.iitk.ac.in).
Jamuna Kanta Sing is with the Department of Computer Science and Engineering, Jadavpur University, Kolkata – 700032, India (e-mail: jksing@ieee.org).

variations and changeable camera viewpoints. Some widely used pattern representation techniques have been introduced like PCA based [15] and LDA-kernel based [16] ear recognitions. In the former technique, Principal Component Analysis has been applied to ear recognition, where the recognition performance is achieved under limited conditions and the later technique uses the combination of Linear discriminant analysis (LDA) and kernel techniques, which overcomes the limitations posed by PCA. However, the problem of generalized eigenvalues is not being solved in the later work [16]. Some feature based and geometric measurement based techniques have been applied to ear recognition successfully in [17], [18]. In [17], authors proposed automatic ear recognition using Voronoi diagram and curve segments. The authors of [18] proposed ear biometrics based on force field transformations. However, the feature based techniques are sensitive and incapable of capturing relevant information from the ear due to variation in pose. Another work proposed in [19], where the authors have applied block-based multi-resolution techniques for ear recognition using wavelet transform and Local Binary Pattern (LBP).

The main aim of the proposed research is to present an invariant feature descriptor based ear recognition that would be useful for considerable pose variations and occlusions. In addition to this it has been determined that while the features are detected from some selected regions where intensity variations are minimum, the detected features are more useful for recognition work rather than the features, which are detected from the whole subject's pattern without color homogeneity consideration, in particular.

Some interesting object recognition works have been proposed and introduced based on scale invariant feature transform (SIFT) [11], [12]. In general object recognition SIFT has been proved to be effective and robust and the uses of SIFT descriptor recently been introduced in different biometrics traits including face [20], fingerprint [21], multimodal biometrics [22].

In this paper, the performance of ear recognition system is further improved based on the object recognition techniques using SIFT features proposed in [12]. Some ear biometric techniques [13], [23] have already been implemented using SIFT features, which demonstrate good results. However, these techniques are still inaccurate for pose variations, non-uniform lighting conditions, background clutter and occlusion. For instance in [24], an ear recognition is introduced using SIFT features, where a person is recognized by the number of keypoints matched and the average distance of the closest square distance. Also, the recognition performance is compared with the two different techniques that use PCA and force field features, respectively. Another work proposed in [25] is basically developed by using invariant SIFT features, while a model-based approach is introduced by capitalizing explicit structure towards robust implementation to noise and occlusion. In [23], an ear recognition is presented, where SIFT features are detected from the multi-pose ear images and fused together by fusion template method. The works discussed in [23], [24], [25] could not deal with the homogeneity in an ear image, which performs with pose variations, occlusion and clutter. An interesting work has proposed in [13], where the authors are tried to address the issues of pose variation, background clutter, occlusion and the ears are taken as planar surface that creates a homography transform using SIFT for registration purpose.

The proposed work presents a robust ear recognition system using Gaussian mixture model (GMM) [27] and SIFT features, where an ear image color is transformed into Gaussian mixture model of pixels. These pixels belong to the densest regions/clusters containing minimum fraction of total pixels in an ear image. In the proposed work takes into consideration not only pose variations, clutter, and occlusion successfully, but also considers the homogeneity with identical color slice regions/clusters, which removes some drawbacks reported earlier [13]. In the subsequent steps, invariant SIFT keypoints are detected only from the color slice regions and fused together for verification. The proposed technique has been tested on IITK ear database. The experimental results demonstrate the effectiveness of the proposed ear system. On the use of combination technique of SIFT features and color homogeneity model is a step towards achieving higher accuracy with 2D ear images.

The paper is organized as follows. GMM color skin model is described for ear images in Section II. Section III briefly discusses the SIFT descriptor. SIFT keypoint features fusion is discussed in Section IV. In Section V, experimental results are demonstrated and finally, the conclusion is made in Section VI.

II. VECTOR QUANTIZED MODELING OF EAR USING GMM

For segmentation of pixels in detected ear image based on the probabilities of identical color spaces, vector quantization is applied [26] to cluster the colors of pixels. Vector quantization can be considered as a fitting model, where the clusters are represented by conditional density functions. In this fitting model, predetermined set of probabilities are the weights. Data contained within vector quantization framework can be fitted with Gaussian mixture models [27] and the probability density function of a dataset is represented as a collection of Gaussians. This convention can be represented by the following equation:

$$f(D) = \sum_{i=1}^{N} P_i f(D \mid i) \qquad (1)$$

where, $N$ is the number of clusters or slice regions in ear image, $p_i$ is the prior probability of cluster $i$ and $f_i(D)$ is the probability density function of cluster $i$. The conditional

probability density function $f_i(D)$ can be represented as

$$f(D|i) = \frac{\exp(-\frac{1}{2}(D-m_i)^t \sum_i^{-1}(D-m_i))}{(2\pi)^{P/2} |\sum_i|^{1/2}} \quad (2)$$

where, $D \in R^P$, and $m_i$ and $\sum_i$ are the mean and covariance matrix of cluster $i$, respectively.

### A. Color Similarity Measurement Using K-L Divergence

Once Gaussian mixture models [27] for color pixels have been formed in the cropped ear images as shown in Fig. 1, K-L divergence theory is used to keep the color consistency in the slice regions independently. K-L divergence [28] is not only used for keep color consistency, but also for finding similarity among the ear images in terms of Gaussian color models [27], [28].

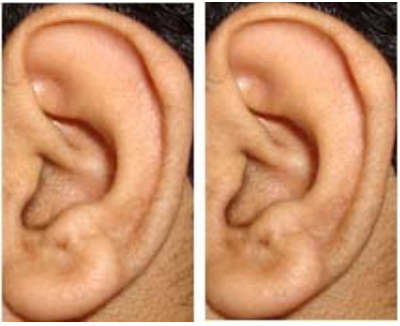

Fig. 1. A pair of ear images is shown.

Generally, Kullback-Leibler divergence or relative entropy is used for measure the dissimilarity between two probability density functions *(PDF) P(D)* and *Q(D)*. K-L divergence can be expressed as follows:

$$KL(P(D) \| Q(D)) = E_{P(D)}\left[\log \frac{P(D)}{Q(D)}\right] \quad (3)$$

The K-L divergence [28] is always nonnegative. K-L divergence would be zero if the two probability density function would be exactly same, otherwise, not.

The approximation of Equation (3) can be written as

$$KL(P(D)\|Q(D)) \cong \sum_{i=1}^{N} P_i \min_j (KL(f_P(D|i)\|f_Q(D|j))) + \log\left[\frac{P_i(D)}{Q_j(D)}\right] \quad (4)$$

where, $f_P(D|i)$ and $f_Q(D|i)$ are multivariate Gaussians [26] in Equation (4).

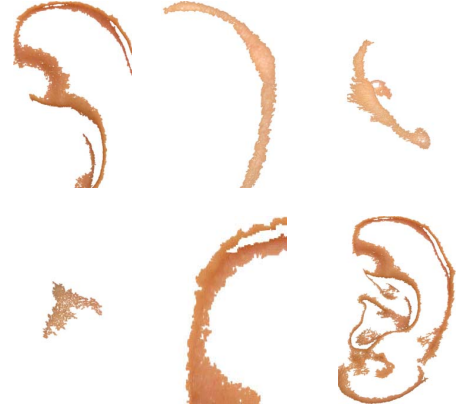

Fig. 2 Color segmented ear slice regions are shown for an ear image

### B. Classification of Ear Skin Color

After modeling ear image using Gaussian mixture model and finding similarity [26], [27] between ear images using K-L divergence [28], the pixels are now classified into several clusters or regions of identical approximated color regions. The algorithm discussed in Section 3 partitions the color pixels within the mask area or within the cropped region of ear image. GMM is used to fit the pixels of unmasked regions in the image. Each slice region is represented by multivariate Gaussian and the weighted collection of multivariate Gaussians approximate the distribution of pixels of cropped ear images. For illustration, see the clustered color slice regions shown in Fig. 2.

In this work, an assumption has made that each ear image can produced different number of slice regions. However, the number of extracted slice regions may have identical with their approximated locations for multiple ear instances of a subject and the total number of detected keypoints may be approximately identical for each color segmented ear image. Let, $n$ is the number of pixels inside the cropped region of ear image and $k_{intra-class}$ is the number of clusters or slice regions, the cluster numbers of a pair of instances for every individual can be represented as

$$k_1, k_2 < n$$
and
$$k_1 < k_{intra-class} < k_2 \quad (5)$$
or
$$k_2 < k_{intra-class} < k_1$$

where, $k_1$ and $k_2$ are the cluster numbers, respectively, for two instances of a subject.

### III. SIFT KEYPOINT EXTRACTION

The Scale Invariant Feature Transform(SIFT) descriptor, has been proposed by [12] and proved to be invariant to image rotation, scaling, partly illumination changes and the

3D camera view. The investigation of SIFT features for biometric authentication has been explored further in [20], [21], [22]. Initially, a pyramid of images is created by convolving the original image by a set of Difference-of-Gaussian (DOG) kernels. The difference of the Gaussian function is increased by a factor in each step. The SIFT descriptor detects feature points efficiently through a staged filtering approach that identifies stable points in the scale-space of the resulting image pyramid. Local feature points are extracted through selecting the candidates for feature points by searching peaks in the scale-space from a DoG function. Further the feature points are localized using the measurement of their stability and assign orientations based on local image properties. Finally, the feature descriptors, which represent local shape distortions and illumination changes, are determined.

In the proposed work, first, the model ear image is normalized by histogram equalization and after normalization SIFT features are extracted from the fused image. Each feature point is composed of four types of information – spatial location ($x$, $y$), scale ($S$), orientation ($\theta$) and Keypoint descriptor ($K$). For the sake experiment, only keypoint descriptor [12], [20] information has been taken which consists of a vector of 128 elements representing neighborhood intensity changes of each keypoint. More formally, local image gradients are measured at the selected scale in the region around each keypoint. The measured gradients information is then transformed into a vector representation that contains a vector of 128 elements for each keypoints calculated over extracted keypoints. These keypoint descriptor vectors represent local shape distortions and illumination changes. In Fig. 3, SIFT features extraction is shown for an ear image.

## IV. FEATURE LEVEL FUSION OF DETECTED SIFT KEYPOINTS AND VERIFICATION

In the proposed ear recognition model, SIFT features detected from color-segmented slice regions are fused together by concatenation. The keypoints are extracted from different slice regions are taken to make an augmented group of features for both the reference ear model and the probe ear model. The proposed fusion strategy uses feature level fusion approach, which is a simple concatenation of the feature sets, obtained from different color segmented slice regions.

Let a segmented ear image $I$ is given, where the independent color slice regions, $S$ are extracted by the method discussed in Section 2. In each slice regions the SIFT feature points are varying. After extraction of the SIFT feature points from the segmented slice regions, the feature points are gathered together by concatenation into an augmented group for each reference model and probe model. Finally, the matching between these two sets of augmented groups are performed using Euclidean distance and nearest neighbor approach. While matching are accomplished between a pair of segmented ear images, the matching scores are obtained and decision for user verification is done by comparing the score against a threshold ( ). In order to obtain fused sets of features for both the reference and the probe models, the keypoints are detected in varying number for each segment region as $K_1$, $K_2$, $K_3$, ……..$K_S$. Now, an augmented set is obtained $DS$ of SIFT features by concatenation as follows

$$DS = \{K_1 \cup K_2 \cup K_3 \cup ..... \cup K_S\} \quad (6)$$

The feature set $DS$ represents the proximity among detected SIFT features of the color slice regions.

Finally, the final matching distance $D_{final}$ ($DS_{probe}$, $DS_{reference}$) is computed on the basis of the number of keypoints paired between two sets of features. The similarity score as follows

$$D_{final} = \sqrt{\sum_{i \in D_{probe}, j \in D_{reference}}(DS_{probe}(K_i) - DS_{reference}(K_j))^2} \leq \psi \quad (7)$$

where, $DS_{probe}$ and $DS_{reference}$ are the concatenated feature sets for both the probe model and the reference model, and is the threshold determined from a subset of database. As for the matching threshold, this ear set is disjoint from the image sets used for testing and validation.

## V. EXPERIMENTAL RESULTS

The experiments with the proposed technique are accomplished on the ear database collected at IIT Kanpur. The database consists of 800 ear images of 400 individuals and all the frontal view ear images are considered for evaluation. The ear images are taken in controlled environment in different sessions. The ear viewpoints are consistently kept neutral and the ear images are downscaled to 237×125 pixels with 500 dpi resolution.

The following protocols are implemented for reference model and probe model selection from the whole database.

**Reference model**: For reference model, single ear image is enrolled for each individual from the set of 800 images. Therefore, 400 images are considered for training session.

**Probe model:** For the probe model set, the remaining 400 images are used for testing and evaluation. Therefore, during matching between pair of reference and probe models is yielding 400 genuine scores and 400×399 imposter scores.

The experiments are conducted in two sessions. In the first session, the ear verification is performed with SIFT features before color segmentation into slice regions and in the next session, verification is performed with the SIFT keypoint features detected from segmented slice regions.

Both the experimental sessions are conducted with the IITK Ear database. From the generated genuine and imposter scores the receiver operating characteristics (ROC) curves are determined and from the ROC curves, true positive rates (TP) and false positive rates (FP) are determined for each respectively. Table I shows the comparative analysis of ear recognition system using SIFT features prior to after color segmentation. The combined ROC curve for different experiments is shown in Fig. 4.

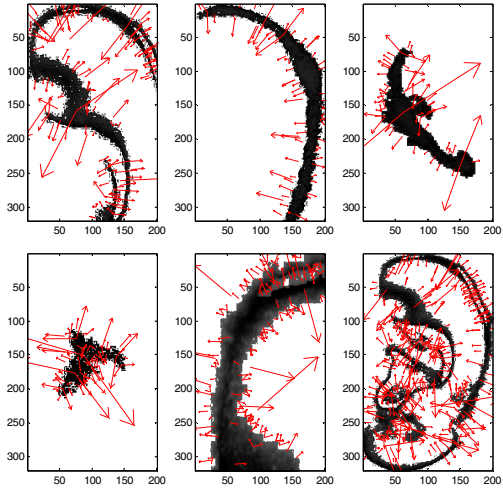

Fig. 3 SIFT keypoint features extraction is shown from the slice regions

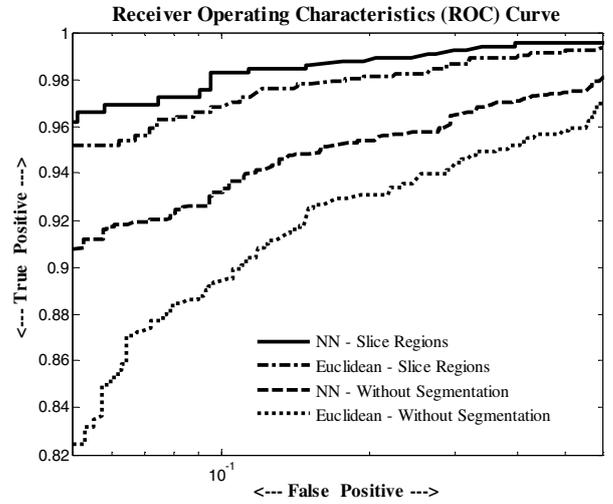

Fig. 4 ROC curves for different ear recognition methods with and without segmentation where NN stands for Nearest Neighbor approach and ED stands for Euclidean Distance approach.

individual experiment by varying the threshold according to the experimental needs.

The accuracy is computed by setting the verification threshold corresponding to minimal value of both (1-TP) and FP in both the sessions. It has been determined that, when Euclidean distance metric is used for verification before color segmentation, the system achieved accuracy of 91.09% with FP and TN of about 9.56% and 8.26% respectively, and when nearest neighbor approach is used for verification before color segmentation, the system generates 93.01% recognition accuracy with FP and TN of about 4.38% and 9.6%, respectively. Due to some false matches from non-segmented slice regions, recognition accuracy often degrades. On the other hand, when the ear image is segmented into some color slice regions, overall system accuracy is increasing radically. False pair SIFT keypoints are often found in non-segmented color slice regions. So, to minimize the false match pairs of SIFT keypoints corresponding to a pair of ear images and increase the true match pairs of keypoints, ear image is segmented into a number of segmented slice regions. These color segmented slice regions represent the area, where most of the true match keypoints are found. When Euclidean distance is used for verification with these segmented slice regions only, the ear recognition system achieved 94.31% recognition accuracy with FP is of 4.22% and TN is of 7.16%, respectively. In another case, when nearest neighbor is used for verification with these slice regions, the system achieved 96.93% recognition accuracy and the computed FP and TN error rates are 2.14% and 4%,

Apart from the robust performance of the proposed system, the results have been compared with the ear recognition systems that are developed using the SIFT features and was proposed by several researchers. There is a SIFT-based ear recognition work [13], which achieves 96% recognition accuracy for baseline recognition. In addition, another set of recognition results have presented by considering some variations in nearby clutter, occlusion and pose changes to ear images. However, the accuracy result achieved in the proposed work gives 96.93%, which outperforms the ear recognition technique discussed in [13], [25]. In [23], the authors have developed an ear recognition system, which uses SIFT descriptor for feature extraction from multi-pose ear images and finally, recognition accomplishes by template fusion. However the proposed method also outperforms the work presented in [23].

Color segmentation of ear images into several color similarity slice regions not only reduce the false matches by discarding non-segmented color regions, but also achieves desired recognition accuracy.

Table I Performance of ear recognition system prior to and after color segmentation. ED refers to Euclidean Distance and NN refers to Nearest Neighborhood approach.

| Approach | Distance | Accuracy | False Positive | True Negative |
|---|---|---|---|---|
| Prior to Color Segmentation | ED | 91.09 | 9.56 | 8.26 |
| | NN | 93.01 | 4.38 | 9.60 |
| After Color | ED | 94.31 | 4.22 | 7.16 |
| Segmentation | NN | 96.93 | 2.14 | 4.00 |

## VI. CONCLUSION

This paper presents an efficient and robust ear recognition system, which uses SIFT descriptor for feature extraction from color similarity clustered regions. It has been estimated that, the segmented regions are the area from where the maximum number of keypoints are found as matching points. The remaining parts of the ear image are discarded, which are not included in color slice regions. The proposed system shows robust performance as the whole area is not considered for verification and also the Gaussian mixture model framework with K-L divergence proves to be a better solution for successfully dividing the ear image into a number of segmented regions. These segmented regions acts as the high probability matching regions for SIFT features. The experiments have been conducted in two different sessions. In the first session, results are obtained prior to segmentation of ear image, the results have been supported with the two distance metrics. In the next session, segmented regions are considered for feature extraction and matching via SIFT keypoint features. During recognition in the second session the system performs with an accuracy of more than 96% using nearest neighborhood approach. This proves that the system can be deployed for high security applications.